\documentclass{article}

\usepackage[nonatbib, final]{nips_2016}

\usepackage[utf8]{inputenc} % allow utf-8 input
\usepackage[T1]{fontenc}    % use 8-bit T1 fonts
\usepackage{hyperref}       % hyperlinks
\usepackage{url}            % simple URL typesetting
\usepackage{booktabs}       % professional-quality tables
\usepackage{amsfonts}       % blackboard math symbols
\usepackage{nicefrac}       % compact symbols for 1/2, etc.
\usepackage{microtype}      % microtypography
\usepackage{graphicx}
\usepackage{amsmath}
\usepackage{subcaption}
\usepackage[ruled,vlined,linesnumbered]{algorithm2e}
\usepackage{setspace}
\usepackage{multirow}
\usepackage{multicol}
\usepackage{color}

\title{Doubly Convolutional Neural Networks}

\author{
  Shuangfei Zhai \\
  Binghamton University\\
  Vestal, NY 13902, USA \\
  \texttt{szhai2@binghamton.edu} \\
  \And
  Yu Cheng \\
  IBM T.J. Watson Research Center \\
  Yorktown Heights, NY 10598, USA \\
  \texttt{chengyu@us.ibm.com} \\
  \And
  Weining Lu \\
  Tsinghua University \\
  Beijing 10084, China \\
  \texttt{luwn14@mails.tsinghua.edu.cn} \\
  \And
  Zhongfei (Mark) Zhang \\
  Binghamton University\\
  Vestal, NY 13902, USA \\
  \texttt{zhongfei@cs.binghamton.edu} \\
}

\begin{document}

\maketitle

\begin{abstract}
Building large models with parameter sharing accounts for most of the success of deep convolutional neural networks (CNNs). In this paper, we propose doubly convolutional neural networks (DCNNs), which significantly improve the performance of CNNs by further exploring this idea. In stead of allocating a set of convolutional filters that are independently learned, a DCNN maintains groups of filters where filters within each group are translated versions of each other. Practically, a DCNN can be easily implemented by a two-step convolution procedure, which is supported by most modern deep learning libraries. We perform extensive experiments on three image classification benchmarks: CIFAR-10, CIFAR-100 and ImageNet, and show that DCNNs consistently outperform other competing architectures. We have also verified that replacing a convolutional layer with a doubly convolutional layer at any depth of a CNN can improve its performance. Moreover, various design choices of DCNNs are demonstrated, which shows that DCNN can serve the dual purpose of building more accurate models and/or reducing the memory footprint without sacrificing the accuracy.
\end{abstract}

\section{Introduction}
In recent years, convolutional neural networks (CNNs) have achieved great success to solve many problems in machine learning and computer vision. CNNs are extremely parameter efficient due to exploring the translation invariant property of images, which is the key to training very deep models without severe overfitting. While considerable progresses have been achieved by aggressively exploring deeper architectures \cite{alexnet,vggnet,inception,residualnet} or novel regularization techniques \cite{dropout,batchnorm} with the standard "convolution + pooling" recipe, we contribute from a different view by providing an alternative to the default convolution module, which can lead to models with even better generalization abilities and/or parameter efficiency. 

\begin{figure}[t]
\centering
\includegraphics[width=0.98\textwidth,height=0.17\textwidth]{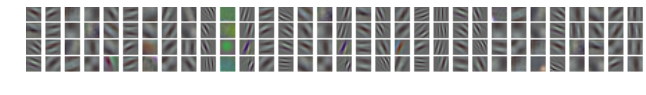}
\caption{Visualization of the $11\times11$ sized first layer filters learned by AlexNet \cite{alexnet}. Each column shows a filter in the first row along with its three most \textit{3-translation-correlated} filters. Only the first 32 filters are shown for brevity.}
\label{fig:filter}
\end{figure}

Our intuition originates from observing well trained CNNs where many of the learned filters are the slightly translated versions of each other. To quantify this in a more formal fashion, we define the \textit{$k$-translation correlation} between two convolutional filters within a same layer $\mathbf{W}_i, \mathbf{W}_j$ as:
\begin{equation} \label{eq:transcorrelation}
\rho_k(\mathbf{W}_i, \mathbf{W}_j) = \max_{x, y \in \{-k, ..., k\},  (x,y) \neq (0, 0)} \frac{<\mathbf{W}_i, T(\mathbf{W}_j, x, y)>_f}{\|\mathbf{W}_i\|_2\|\mathbf{W}_j\|_2},
\end{equation}
where $T(\cdot, x, y)$ denotes the translation of the first operand by $(x,y)$ along its spatial dimensions, with proper zero padding at borders to maintain the shape; $<\cdot, \cdot>_f$ denotes the flattened inner product, where the two operands are flattened into column vectors before taking the standard inner product; $\|\cdot\|_2$ denotes the $\ell_2$ norm of its flattened operand. In other words, the \textit{$k$-translation correlation} between a pair of filters indicates the maximum correlation achieved by translating one filter up to $k$ steps along any spatial dimension. As a concrete example, Figure \ref{fig:filter} demonstrates the \textit{$3$-translation correlation} of the first layer filters learned by the AlexNet \cite{alexnet}, with the weights obtained from the Caffe model zoo \cite{caffe}. In each column, we show a filter in the first row and its three most \textit{$3$-translation-correlated} filters (that is, filters with the highest \textit{$3$-translation correlations}) in the second to fourth row. Only the first 32 filters are shown for brevity. It is interesting to see for most filters, there exist several filters that are roughly its translated versions.

\begin{figure}[t]
\centering
\includegraphics[scale=0.25]{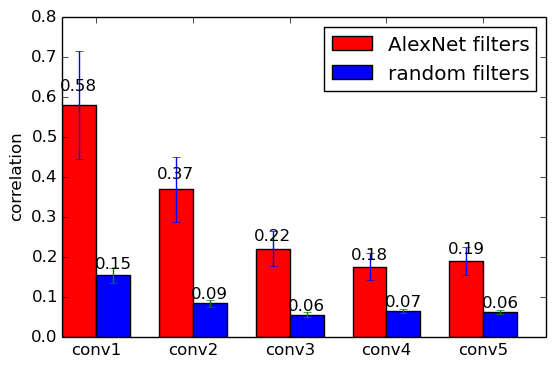}
\includegraphics[scale=0.25]{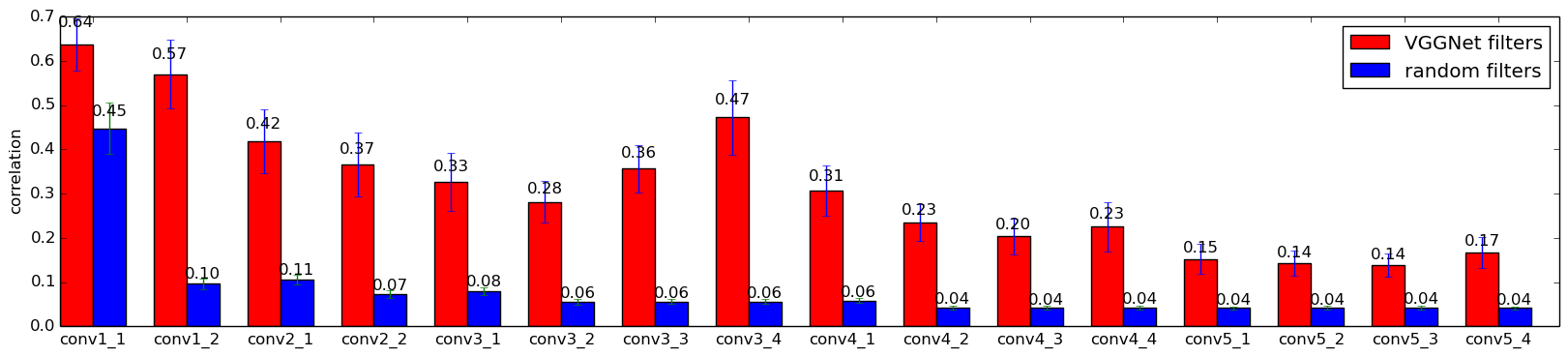}
\caption{Illustration of the \textit{averaged maximum 1-translation correlation}, together with the standard deviation, of each convolutional layer for AlexNet \cite{alexnet} (left), and the 19-layer VGGNet \cite{vggnet} (right), respectively. For comparison, for each convolutional layer in each network, we generate a filter set with the same shape from the standard Gaussian distribution (the blue bars). For both networks, all the convolutional layers have \textit{averaged maximum 1-translation correlations} that are significantly larger than their random counterparts.}
\label{fig:alllayers}
\end{figure}

In addition to the convenient visualization of the first layers, we further study this property at higher layers and/or in deeper models. To this end, we define the \textit{averaged maximum $k$-translation correlation} of a layer $\mathbf{W}$ as $\bar{\rho}_k(\mathbf{W}) = \frac{1}{N}\sum_{i=1}^{N} \max_{j=1, j \neq i}^N {\rho_k(\mathbf{W}_i, \mathbf{W}_j)}$, where $N$ is the number of filters. Intuitively, the $\bar{\rho}_k$ of a convolutional layer characterizes the average level of translation correlation among the filters within it. We then load the weights of all the convolutional layers of AlexNet as well as the 19-layer VGGNet \cite{vggnet} from the Caffe model zoo, and report the \textit{averaged maximum $1$-translation correlation} of each layer in Figure \ref{fig:alllayers}. In each graph, the height of the red bars indicates the $\bar{\rho}_1$ calculated with the weights of the corresponding layer. As a comparison, for each layer we have also generated a filter bank with the same shape but filled with standard Gaussian samples, whose $\bar{\rho}_1$ are shown as the blue bars. We clearly see that all the layers in both models demonstrate averaged maximum translation correlations that are significantly higher than their random counterparts. In addition, it appears that lower convolutional layers generally have higher translation correlations, although this does not strictly hold (e.g., conv3\_4 in VGGNet).

Motivated by the evidence shown above, we propose the doubly convolutional layer (with the double convolution operation), which can be plugged in place of a convolutional layer in CNNs, yielding the doubly convolutional neural networks (DCNNs). The idea of double convolution is to learn groups filters where filters within each group are translated versions of each other. To achieve this, a doubly convolutional layer allocates a set of \textit{meta filters} which has filter sizes that are larger than the effective filter size. Effective filters can be then extracted from each meta filter, which corresponds to convolving the meta filters with an identity kernel. All the extracted filters are then concatenated, and convolved with the input. Optionally, one can also choose to pool along activations produced by filters from the same meta filter, in a similar spirit to the maxout networks \cite{maxout}. We also show that double convolution can be easily implemented with available deep learning libraries by utilizing the efficient convolutional kernel. In our experiments, we show that the additional level of parameter sharing by double convolution allows one to build DCNNs that yield an excellent performance on several popular image classification benchmarks, consistently outperforming all the competing architectures with a margin. We have also confirmed that replacing a convolutional layer with a doubly convolutional layer consistently improves the performance, regardless of the depth of the layer. Last but not least, we show that one is able to balance the trade off between performance and parameter efficiency by leveraging the architecture of a DCNN.

\section{Model}
\subsection{Convolution}
We define an image $\mathcal{I} \in R^{c \times w \times h}$ as a real-valued 3D tensor, where $c$ is the number of channels; $w,h$ are the width and height, respectively. We define the convolution operation, denoted by $\mathcal{I}^{\ell + 1} = \mathcal{I}^{\ell} \ast \mathbf{W}^{\ell}$, as follows:
\begin{equation}
\label{eq:conv}
\begin{split}
&\mathcal{I}^{\ell + 1}_{k,i,j} = \sum_{c'\in[1,c], i' \in [1,z], j' \in [1,z]}{\mathbf{W}^{\ell}_{k,c',i',j'} \mathcal{I}^{\ell}_{c', i+i'-1, j+j'-1}}, \\
& k \in [1, c^{\ell + 1}], i \in [1, w^{\ell + 1}], j \in [1, h^{\ell + 1}].
\end{split}
\end{equation}
Here $\mathcal{I}^{\ell}\in R^{c^{\ell} \times w^{\ell} \times h^{\ell}}$ is the input image; $\mathbf{W}^{\ell} \in R^{c^{\ell +1}\times c^{\ell} \times z \times z}$ is a set of $c^{\ell +1}$ filters, with each filter of shape $c^{\ell} \times z \times z$; $\mathcal{I}^{\ell + 1}\in R^{c^{\ell + 1} \times w^{\ell + 1} \times h^{\ell + 1}}$ is the output image. The spatial dimensions of the output image $w^{\ell + 1}, h^{\ell + 1}$ are by default $w^{\ell} + z - 1$ and $h^{\ell} + z - 1$, respectively (aka, valid convolution), but one can also pad a number of zeros at the borders of $\mathcal{I}^{\ell}$ to achieve different output spatial dimensions (e.g., keeping the spatial dimensions unchanged). In this paper, we use a loose notation by freely allowing both the LHS and RHS of $\ast$ to be either a single image (filter) or a set of images (filters), with proper convolution along the non-spatial dimensions.

A convolutional layer can thus be implemented with a convolution operation followed by a nonlinearity function such as ReLU, and a convolutional neural network (CNN) is constructed by interweaving several convolutoinal and spatial pooling layers. 
% The most distinct property of a CNN is that the filters are shared across locations, which means that the number of parameters does not grow with the size of the input image. In such way, one is able to build very large models without the risk of overfitting. 

\subsection{Double convolution} \label{sec:doubleconv} 
% \begin{figure}[t]
% \centering
% \includegraphics[scale=0.3]{filters.png}
% \caption{Visualization of the first layer filters learned by AlexNet \cite{alexnet}.}
% \label{fig:alexnet}
% \end{figure}

\begin{figure}
\begin{subfigure}{0.42\textwidth}
\includegraphics[scale=0.05]{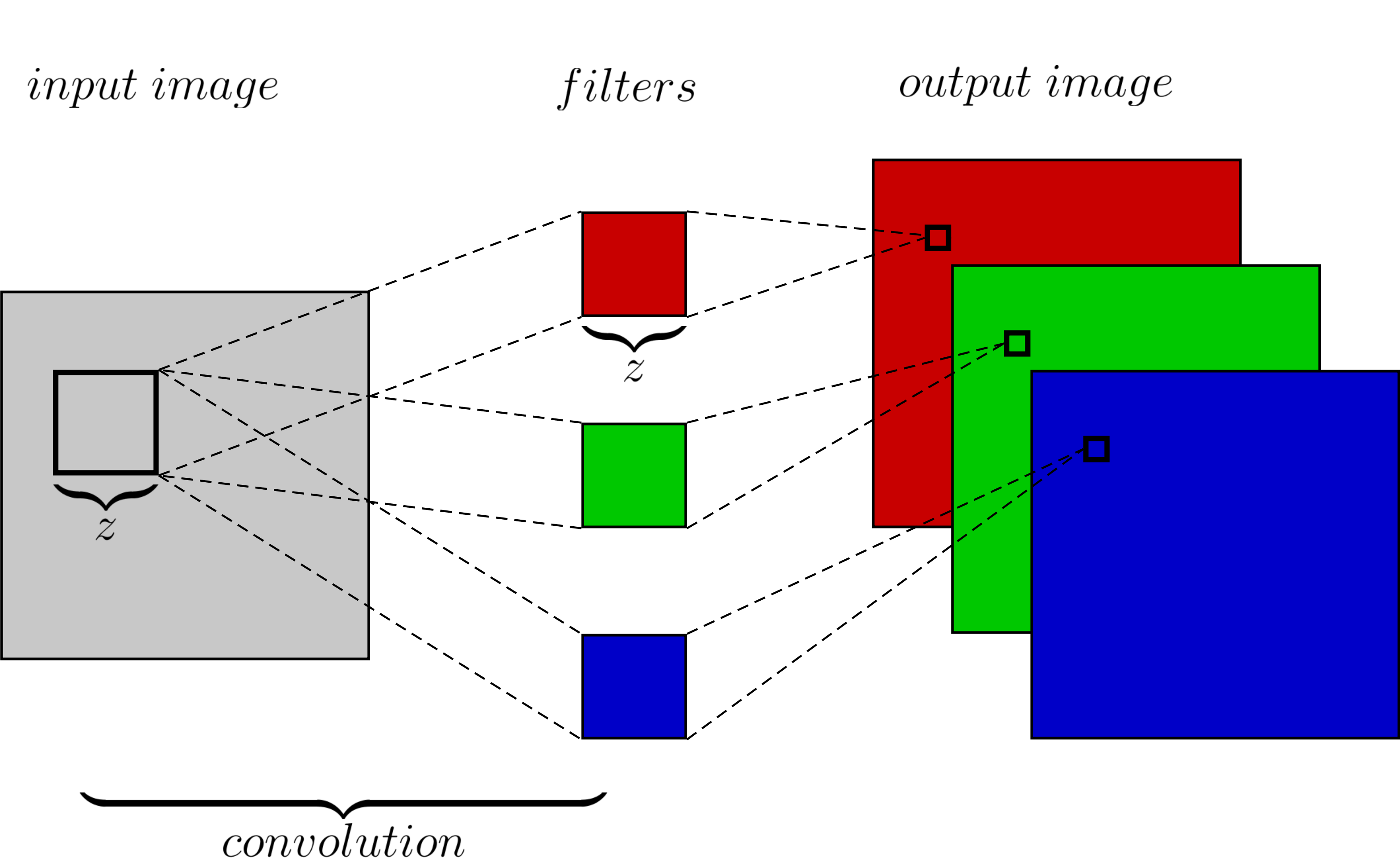}
\end{subfigure}
\begin{subfigure}{0.58\textwidth}
\centering
\includegraphics[scale=0.05]{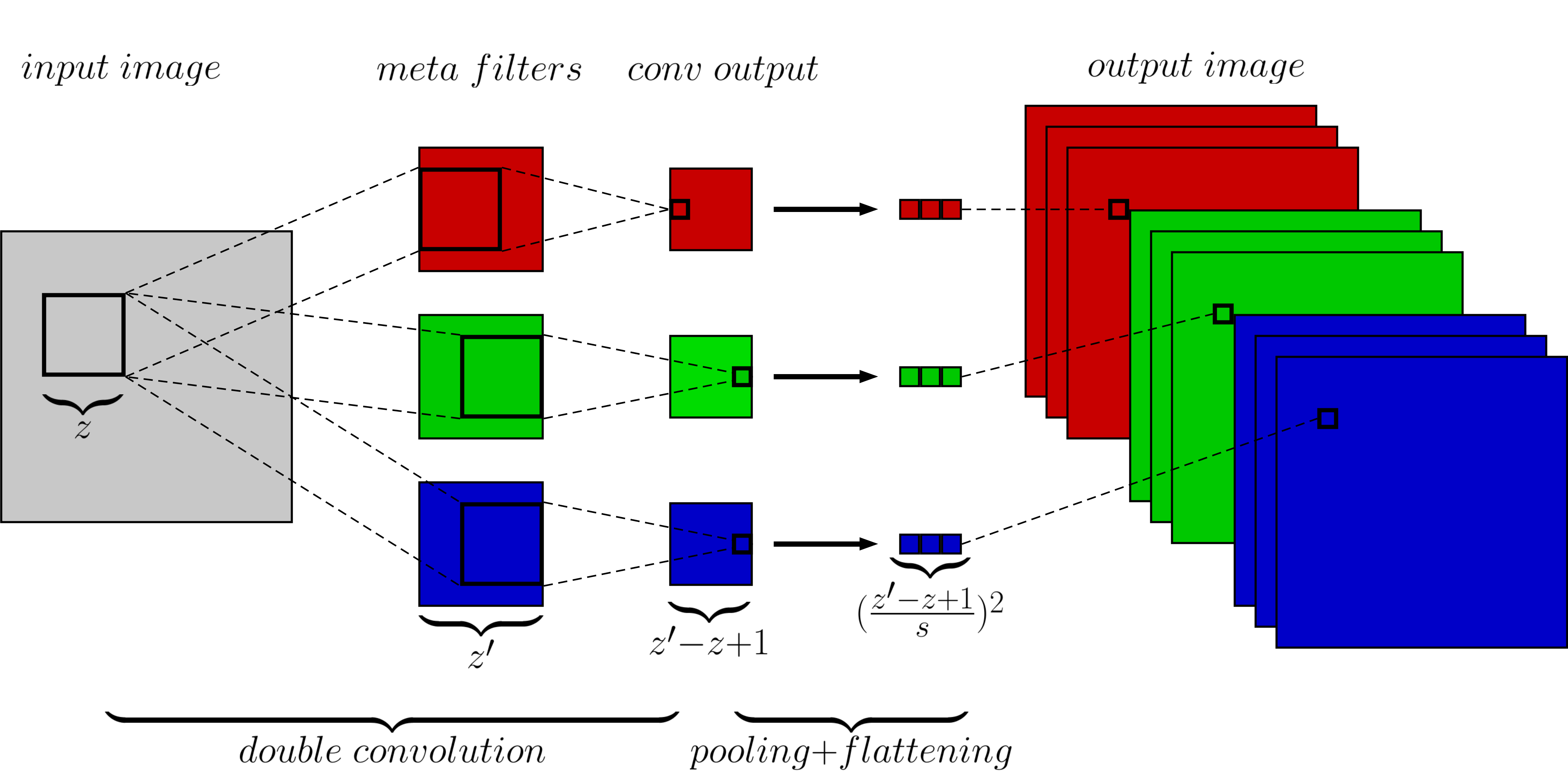}
\end{subfigure}
\caption{The architecture of a convolutional layer (left) and a doubly convolutional layer (right). A doubly convolutional layer maintains meta filters whose spatial size $z' \times z'$ is larger than the effective filter size $z \times z$. By pooling and flattening the convolution output, a doubly convolutional layer produces $(\frac{z' - z + 1}{s})^2$ times more channels for the output image, with $s\times s$ being the pooling size.}
\label{fig:conv}
\end{figure}
We next introduce and define the double convolution operation, denoted by $\mathcal{I}^{\ell + 1} = \mathcal{I}^{\ell} \otimes \mathbf{W}^{\ell}$, as follows: 
\begin{equation}
\begin{split}
&\mathcal{O}^{\ell + 1}_{i,j, k} = \mathbf{W}^{\ell}_k \ast \mathcal{I}^{\ell}_{:, i:(i+z-1),j:(j+z-1)}, \\
&\mathcal{I}^{\ell + 1}_{(nk + 1):n(k+1),i,j} = pool_s(\mathcal{O}^{\ell + 1}_{i,j, k}), n = (\frac{z' - z + 1}{s})^2,\\
&  k \in [1, c^{\ell + 1}], i \in [1, w^{\ell + 1}], j \in [1, h^{\ell + 1}].
\end{split}
\label{eq:doublconv}
\end{equation}
Here $\mathcal{I}^{\ell} \in R^{c^{\ell} \times w^{\ell} \times h^{\ell}}$ and $\mathcal{I}^{\ell + 1} \in R^{nc^{\ell + 1} \times w^{\ell + 1} \times h^{\ell + 1}}$ are the input and output image, respectively. $\mathbf{W}^{\ell} \in R^{c^{\ell + 1}\times c^{\ell} \times z' \times z'}$ are a set of $c^{\ell + 1}$ \textit{meta filters}, with filter size $z' \times z', z' > z$; $\mathcal{O}^{\ell + 1}_{i,j, k} \in R^{(z'-z+1)\times(z'-z+1)}$ is the intermediate output of double convolution; $pool_s(\cdot)$ defines a spatial pooling function with pooling size $s \times s$ (and optionally reshaping the output to a column vector, inferred from the context); $\ast$ is the convolution operator defined previously in Equation \ref{eq:conv}. 

In words, a double convolution applies a set of $c^{\ell + 1}$ meta filters with spatial dimensions $z' \times z'$, which are larger than the effective filter size $z \times z$. Image patches of size $z \times z$ at each location $(i,j)$ of the input image, denoted by $\mathcal{I}^{\ell}_{:, i:(i+z-1),j:(j+z-1)}$, are then convolved with each meta filter, resulting an output of size $z' - z + 1 \times z' - z + 1$, for each $(i,j)$. A spatial pooling of size $s \times s$ is then applied along this resulting output map, whose output is flattened into a column vector. This produces an output feature map with $nc^{\ell + 1}$ channels. The above procedure can be viewed as a two step convolution, where image patches are first convolved with meta filters, and the meta filters  then slide across and convolve with the image, hence the name double convolution.

A doubly convolutional layer is by analogy defined as a double convolution followed by a nonlinearity; and substituting the convolutional layers in a CNN with doubly convolutional layers yields a doubly convolutional neural network (DCNN). In Figure \ref{fig:conv} we have illustrated the difference between a convolutional layer and a doubly convolutional layer. It is possible to vary the combination of $z, z', s$ for each doubly convolutional layer of a DCNN to yield different variants, among which three extreme cases are:

(1) \textbf{CNN}: Setting $z' = z$ recovers the standard CNN; hence, DCNN is a generalization of CNN.

(2) \textbf{ConcatDCNN}: Setting $s=1$ produces a DCNN variant that is maximally parameter efficient. This corresponds to extracting all sub-regions of size $z \times z$ from a $z' \times z'$ sized meta filter, which are then stacked to form a set of $(z' - z + 1)^2$ filters with size $z \times z$. With the same amount of parameters, this produces $\frac{(z' - z +1)^2 z^2}{(z')^2}$ times more channels for a single layer.

(3) \textbf{MaxoutDCNN}: Setting $s = z' - z + 1$, i.e., applying global pooling on $\mathcal{O}^{\ell+1}$, produces a DCNN variant where the output image channel size is equal to the number of the meta filters. Interestingly, this yields a parameter efficient implementation of the maxout network \cite{maxout}. To be concrete, the maxout units in a maxout network are equivalent to pooling along the channel (feature) dimension, where each channel corresponds to a distinct filter. MaxoutDCNN, on the other hand, pools along channels which are produced by the filters that are translated versions of each other. Besides the obvious advantage of reducing the number of parameters required, this also acts as an effective regularizer, which is verified later in the experiments at Section \ref{sec:exp}.

Implementing a double convolution is also readily supported by most main stream GPU-compatible deep learning libraries (e.g., Theano which is used in our experiments), which we have summarized in Algorithm \ref{alg:doubleconv}. In particular, we are able to perform double convolution by two steps of convolution, corresponding to line 4 and line 6, together with proper reshaping and pooling operations. The first convolution extracts overlapping patches of size $z \times z$ from the meta filters, which are then convolved with the input image. Although it is possible to further reduce the time complexity by designing a specialized double convolution module, we find that Algorithm \ref{alg:doubleconv} scales well to deep DCNNs, and large datasets such as ImageNet.

\begin{algorithm}[h]
% \setstretch{1.}'
\small
\KwIn{Input image $\mathcal{I}^{\ell} \in R^{c^{\ell} \times w^{\ell} \times h^{\ell}}$, meta filters $\mathbf{W}^{\ell} \in R^{c^{\ell + 1} \times z' \times z'}$, effective filter size $z \times z$, pooling size $s \times s$.}
\KwOut{Output image $\mathcal{I}^{\ell + 1} \in R^{nc^{\ell + 1} \times w^{\ell + 1} \times h^{\ell + 1}}$, with $n = \frac{(z' - z +1)^2}{s^2}$.}

\Begin{
$\mathbf{I}^{\ell} \leftarrow \mathit{IdentityMatrix}(c^{\ell}z^2)$ \;
Reorganize $\mathbf{I}^{\ell}$ to shape $c^{\ell}z^2 \times c^{\ell} \times z \times z$\;
$\mathbf{\tilde{W}}^{\ell} \leftarrow \mathbf{W}^{\ell} \ast \mathbf{I}^{\ell}$ \tcc*{output shape: $c^{\ell + 1} \times c^{\ell}z^2 \times (z' - z + 1) \times (z' -z + 1)$} 
Reorganize $\mathbf{\tilde{W}}^{\ell}$ to shape $c^{\ell + 1}(z' - z + 1)^2 \times c^{\ell} \times z \times z$\;
$\mathcal{O}^{\ell + 1} \leftarrow \mathcal{I}^{\ell} \ast \mathbf{\tilde{W}}^{\ell}$ \tcc*{output shape: $c^{\ell + 1}(z'-z+1)^2 \times w^{\ell + 1}\times h^{\ell + 1}$}
Reorganize $\mathcal{O}^{\ell + 1}$ to shape $c^{\ell + 1}w^{\ell + 1}h^{\ell + 1} \times (z' - z + 1) \times (z' - z + 1)$ \;
$\mathcal{I}^{\ell + 1} \leftarrow pool_s(\mathcal{O}^{\ell + 1})$ \tcc*{output shape: $c^{\ell + 1}w^{\ell + 1}h^{\ell + 1} \times \frac{z' - z + 1}{s} \times \frac{z' - z + 1}{s}$}
Reorganize $\mathcal{I}^{\ell + 1}$ to shape $c^{\ell + 1}(\frac{z' - z + 1}{s})^2 \times w^{\ell + 1} \times h^{\ell + 1}$ \;
}
\caption{{\bf Implementation of double convolution with convolution.} }
\label{alg:doubleconv}
\end{algorithm}

\section{Related work}
The spirit of DCNNs is to further push the idea of parameter sharing of the convolutional layers, which is shared by several recent efforts. \cite{cyclic} explores the rotation symmetry of certain classes of images, and hence proposes to rotate each filter (or alternatively, the input) by a multiplication of $90^{\circ}$ which produces four times filters with the same amount of parameters for a single layer. \cite{crelu} observes that filters learned by ReLU CNNs often contain pairs with opposite phases in the lower layers. The authors accordingly propose the concatenated ReLU where the linear activations are concatenated with their negations and then passed to ReLU, which effectively doubles the number of filters. \cite{dilate} proposes the dilated convolutions, where additional filters with larger sizes are generated by dilating the base convolutional filters, which is shown to be effective in dense prediction tasks such as image segmentation. \cite{mba} proposes a multi-bias activation scheme where $k, k \leq 1$, bias terms are learned for each filter, which produces a $k$ times channel size for the convolution output. Additionally, \cite{deepsymmetry,cohen2016group} have investigated the combination of more than one transformations of filters, such as rotation, flipping and distortion. 
Note that all the aforementioned approaches are orthogonal to DCNNs and can theoretically be combined in a single model. The need of correlated filters in CNNs is also studied in \cite{topographic}, where similar filters are explicitly learned and grouped with a group sparsity penalty.

While DCNNs are designed with better performance and generalization ability in mind, they are also closely related to the thread of work on parameter reduction in deep neural networks. The work of Vikas and Tara \cite{NIPS2015_5869} addresses the problem of compressing deep networks by applying structured transforms. \cite{Yu2015} exploits the redundancy in the parametrization of deep architectures by imposing a circulant structure on the projection matrix, while allowing the use of FFT for faster computations. \cite{Yang2015} attempts to obtain the compression of the fully-connected layers of the AlexNet-type network with the Fastfood method. Novikov \textit{et al.} \cite{novikov15tensornet} use a multi-linear transform (Tensor-Train decomposition) to attain reduction of the number of parameters in the linear layers of CNNs. These work differ from DCNNs as most of their focuses are on the fully connected layers, which often accounts for most of the memory consumption. DCNNs, on the other hand, apply directly to the convolutional layers, which provides a complementary view to the same problem.

\section{Experiments}\label{sec:exp}
\subsection{Datasets} \label{sec:dataset}

We conduct several sets of experiments with DCNN on three image classification benchmarks: CIFAR-10, CIFAR-100, and ImageNet. CIFAR-10 and CIFAR-100 both contain 50,000 training and 10,000 testing $32\times32$ sized RGB images, evenly drawn from 10 and 100 classes, respectively. ImageNet is the dataset used in the ILSVRC-2012 challenge, which consists of about 1.2 million images for training and 50,000 images for validation, sampled from 1,000 classes.

\begin{table*}[t]
\scriptsize
\caption{The configurations of the models used in Section
\ref{sec:effective}. The architectures on the CIFAR-10 and CIFAR-100 datasets are the same, except for the top softmax layer (left). The architectures on the ImageNet dataset are variants of the 16-layer VGGNet \cite{vggnet} (right). See the details about the naming convention in Section \ref{sec:architectures}.}
\label{tab:architecture}
\begin{minipage}{0.5\linewidth}
\centering
\begin{tabular}{*3c}
\toprule
CNN & DCNN & MaxoutCNN \\
\midrule
C-128-3&  DC-128-4-3-2 & MC-512-3-4 \\
C-128-3&  DC-128-4-3-2 & MC-512-3-4 \\
\midrule
\multicolumn{3}{c}{P-2} \\
\midrule
C-128-3&  DC-128-4-3-2 & MC-512-3-4 \\
C-128-3&  DC-128-4-3-2 & MC-512-3-4 \\
\midrule
\multicolumn{3}{c}{P-2} \\
\midrule
C-128-3&  DC-128-4-3-2 & MC-512-3-4 \\
C-128-3&  DC-128-4-3-2 & MC-512-3-4 \\
\midrule
\multicolumn{3}{c}{P-2} \\
\midrule
C-128-3&  DC-128-4-3-2 & MC-512-3-4 \\
C-128-3&  DC-128-4-3-2 & MC-512-3-4 \\
\midrule
\multicolumn{3}{c}{P-2} \\
\midrule
\multicolumn{3}{c}{Global Average Pooling} \\
\midrule
\multicolumn{3}{c}{Softmax} \\
\bottomrule
\end{tabular}
\end{minipage}
\begin{minipage}{0.5\linewidth}
\centering
\begin{tabular}{*3c}
\toprule
CNN & DCNN & MaxoutCNN  \\
\midrule
C-64-3&  DC-64-4-3-2 & MC-256-3-4\\
C-64-3&  DC-64-4-3-2 & MC-256-3-4\\
\midrule
\multicolumn{3}{c}{P-2} \\
\midrule
C-128-3&  DC-128-4-3-2 & MC-512-3-4 \\
C-128-3&  DC-128-4-3-2 & MC-512-3-4 \\
\midrule
\multicolumn{3}{c}{P-2} \\
\midrule
C-256-3&  DC-256-4-3-2 & MC-1024-3-4 \\
C-256-3&  DC-256-4-3-2 & MC-1024-3-4 \\
C-256-3&  DC-256-4-3-2 & MC-1024-3-4 \\
\midrule
\multicolumn{3}{c}{P-2} \\
\midrule
C-512-3&  DC-512-4-3-2 & MC-2048-3-4 \\
C-512-3&  DC-512-4-3-2 & MC-2048-3-4 \\
C-512-3&  DC-512-4-3-2 & MC-2048-3-4 \\
\midrule
\multicolumn{3}{c}{P-2} \\
\midrule
C-512-3&  DC-512-4-3-2 & MC-2048-3-4 \\
C-512-3&  DC-512-4-3-2 & MC-2048-3-4 \\
C-512-3&  DC-512-4-3-2 & MC-2048-3-4 \\
\midrule
\multicolumn{3}{c}{P-2} \\
\midrule
\multicolumn{3}{c}{Global Average Pooling} \\
\midrule
\multicolumn{3}{c}{Softmax} \\
\bottomrule
\end{tabular}
\end{minipage}
\end{table*}

\subsection{Is DCNN an effective architecture?} \label{sec:effective}
\subsubsection{Model specifications} \label{sec:architectures}
In the first set of experiments, we study the effectiveness of DCNN compared with two different CNN designs. The three types of architectures subject to evaluation are:

(1) \textbf{CNN}: This corresponds to models using the standard convolutional layers. A convolutional layer is denoted as C-<$c$>-<$z$>, where $c$, $z$ are the number of filters and the filter size, respectively.

(2) \textbf{MaxoutCNN}: This corresponds to the maxout convolutional networks \cite{maxout}, which uses the maxout unit to pool along the channel (feature) dimensions with a stride $k$. A maxout convolutional layer is denoted as MC-<$c$>-<$z$>-<$k$>, where $c$, $z$, $k$ are the number of filters, the filter size, and the feature pooling stride, respectively.

(3) \textbf{DCNN}: This corresponds to using the doubly convolutional layers. We denote a doubly convolutional layer with $c$ filters as DC-<$c$>-<$z'$>-<$z$>-<$s$>, where $z', z, s$ are the meta filter size, effective filter size and pooling size, respectively, as in Equation \ref{eq:doublconv}. In this set of experiments, we use the MaxoutDCNN variant, whose layers are readily represented as DC-<$c$>-<$z'$>-<$z$>-<$z' - z +1$>.

We denote a spatial max pooling layer as P-<$s$> with $s$ as the pooling size. For all the models, we apply batch normalization \cite{batchnorm} immediately after each convolution layer, after which ReLU is used as the nonlinearity (including MaxoutCNN, which makes out implementation slightly different from \cite{maxout}). Our model design is similar to VGGNet \cite{vggnet} where $3\times3$ filter sizes are used, as well as Network in Network \cite{nin} where fully connected layers are completely eliminated. Zero padding is used before each convolutional layer to maintain the spatial dimensions unchanged after convolution. Dropout is applied after each pooling layer. Global average pooling is applied on top of the last convolutional layer, which is fed to a Softmax layer with a proper number of outputs.

All the three models on each dataset are of the same architecture w.r.t. the number of layers and the number of units per layer. The only difference thus resides in the choice of the convolutional layers. Note that the architecture we have used on the ImageNet dataset resembles the 16-layer VGGNet \cite{vggnet}, but without the fully connected layers. The full specification of the model architectures is shown in Table \ref{tab:architecture}.

\subsubsection{Training protocols} \label{sec:training}
We preprocess all the datasets by extracting the mean for each pixel and each channel, calculated on the training sets. All the models are trained with Adadelta \cite{adadelta} on NVIDIA K40 GPUs. Bath size is set as 200 for CIFAR-10 and CIFAR-100, and 128 for ImageNet.

Data augmentation has also been explored. On CIFAR-10 and CIFAR-100, We follow the simple data augmentation as in \cite{vggnet}. For training, 4 pixels are padded on each side of the images, from which $32\times32$ crops are sampled with random horizontal flipping. For testing, only the original $32\times32$ images are used. On ImageNet, $224\times224$ crops are sampled with random horizontal flipping; the standard color augmentation and the 10-crop testing are also applied as in AlexNet \cite{alexnet}.

\subsubsection{Results}
The test errors are summarized in Table \ref{tab:cifar:result} and Table \ref{tab:imagenet:result}, where the relative \# parameters of DCNN and MaxoutCNN compared with the standard CNN are also shown. On the moderately-sized datasets CIFAR-10 and CIFAR-100, DCNN achieves the best results of the three control experiments, with and without data augmentation. Notably, DCNN consistently improves over the standard CNN with a margin. More remarkably, DCNN also consistently outperforms MaxoutCNN, with 2.25 times less parameters. This on the one hand proves that the doubly convolutional layers greatly improves the model capacity, and on the other hand verifies our hypothesis that the parameter sharing introduced by double convolution indeed acts as a very effective regularizer. The results achieved by DCNN on the two datasets are also among the best published results compared with \cite{nin,dsn,apl,elu}.

Besides, we also note that DCNN does not have difficulty scaling up to a large dataset as ImageNet, where consistent performance gains over the other baseline architectures are again observed. Compared with the results of the 16-layer VGGNet in \cite{vggnet} with multiscale evaluation, our DCNN implementation achieves comparable results, with significantly less parameters.

\begin{table*}[h]
\footnotesize
\caption{Test errors on CIFAR-10 and CIFAR-100 with and 
without data augmentation, together with the relative \# parameters compared with the standard CNN.}
\label{tab:cifar:result}
\centering
\begin{tabular}{cccccc}
\toprule
\multirow{2}{*}{\textbf{Model}} & \multirow{2}{*}{\textbf{\# Parameters}} & \multicolumn{2}{c}{\textbf{Without Data Augmentation}} & \multicolumn{2}{c}{\textbf{With Data Augmentation}} \\
\cline{3-6}
& & CIFAR-10 & CIFAR-100 & CIFAR-10 & CIFAR-100 \\
\midrule
CNN & 1. & 9.85\% & 34.26\% & 9.59\% & 33.04\%  \\ 
MaxoutCNN & 4. & 9.56\% & 33.52\% & 9.23\% & 32.37\% \\
DCNN & 1.78 & \textbf{8.58\%} & \textbf{30.35\%} &\bf  7.24\% & \bf 26.53\%  \\
\midrule
NIN \cite{nin} & 0.92 & 10.41\% &35.68\% & 8.81\% & -\\
DSN \cite{dsn} & - & 9.78\% & 34.57\% & 8.22\% & - \\
APL \cite{apl} & - & 9.59\% & 34.40\% & 7.51\% & 30.83\%  \\
ELU \cite{elu} & - & - & - & \textbf{6.55\%} & \textbf{24.28\%}   \\
\bottomrule
\end{tabular}
\end{table*}

\begin{table*}
\begin{center}
 \caption{Test errors on ImageNet, evaluated on the validation set, together with the relative \# parameters compared with the standard CNN.}
 \label{tab:imagenet:result}
 \begin{tabular}{cccc}
 \toprule
 \textbf{Model}  & \textbf{Top-5 Error}  & \textbf{Top-1 Error} & \textbf{\# Parameters} \\ \hline
 CNN  & 10.59\% & 29.42\% & 1.\\ 
 MaxoutCNN  & 9.82\% & 28.4\% & 4.\\ 
 DCNN & \bf 8.23\% & \bf 26.27 \%  & 1.78\\ 
 \midrule
 VGG-16 \cite{vggnet} &7.5\% &24.8\% & 9.3\\
 ResNet-152 \cite{residualnet}& \bf 5.71\% & \bf 21.43\%  & 4.1\\
 GoogLeNet \cite{inception} & 7.9\% & - & 0.47\\ 
 \bottomrule
 \end{tabular}
 \end{center}
 \end{table*}

\subsection{Does double convolution contribute to every layer?}
In the next set of experiments, we study the effect of applying double convolution to layers at various depths. To this end, we replace the convolutional layers at each level of the standard CNN defined in \ref{sec:architectures} with a doubly convolutional layer counterpart (e.g., replacing a C-128-3 layer with a DC-128-4-3-2 layer). We hence define DCNN[i-j] as the network resulted from replacing the $i-j$th convolutional layer of a CNN with its doubly convolutional layer counterpart, and train \{DCNN[1-2], DCNN[3-4], DCNN[5-6], DCNN[7-8]\} on CIFAR-10 and CIFAR-100 following the same protocol as that in Section \ref{sec:training}. The results are shown in Table \ref{tab:layer}. Interestingly, the doubly convolutional layer is able to consistently improve the performance over that of the standard CNN regardless of the depth with which it is plugged in. Also, it seems that applying double convolution at lower layers contributes more to the performance, which is consistent with the trend of translation correlation observed in Figure \ref{fig:alllayers}.

\begin{table*}[t]
\begin{center}
\caption{Inserting the doubly convolutional layer at different depths of the network.}
\label{tab:layer}
\begin{tabular}{cc c}
\toprule
\textbf{Model}  & \textbf{CIFAR-10}  & \textbf{CIFAR-100}  \\ \midrule
 CNN & 9.85\% & 34.26\% \\
 DCNN[1-2] & 9.12\%  &  32.91\% \\ 
 DCNN[3-4]  & 9.23\%  &  33.27\% \\ 
 DCNN[5-6] & 9.45\%  & 33.58\%  \\ 
 DCNN[7-8] & 9.57\%  & 33.72\%  \\ 
 DCNN[1-8] & 8.58\% & 30.35\% \\
 \bottomrule
 \end{tabular}
 \end{center}
 \end{table*}

\subsection{Performance \textit{vs.} parameter efficiency}
In the last set of experiments, we study the behavior of DCNNs under various combinations of its hyper-parameters, $z', z, s$. To this end, we train three more DCNNs on CIFAR-10 and CIFAR-100, namely \{DCNN-32-6-3-2, DCNN-16-6-3-1, DCNN-4-10-3-1\}. Here we have overloaded the notation for a doubly convolutional layer to denote a DCNN which contains correspondingly shaped doubly convolutional layers (the DCNN in Table \ref{tab:architecture} thus corresponds to DCNN-128-4-3-2). In particular, DCNN-32-6-3-2 produces a DCNN with the exact same shape and number of parameters of those of the reference CNN; DCNN-16-6-3-1, DCNN-4-10-3-1 are two ConcatDCNN instances from Section \ref{sec:doubleconv}, which produce larger sized models with same or less amount of parameters. The results, together with the effective layer size and the relative number of parameters, are listed in Table \ref{tab:structure}. We see that all the variants of DCNN consistently outperform the standard CNN, even when fewer parameters are used (DCNN-4-10-3-1). This verifies that DCNN is a flexible framework which allows one to either maximize the performance with a fixed memory budget, or on the other hand, minimize the memory footprint without sacrificing the accuracy. One can choose the best suitable architecture of a DCNN by balancing the trade off between performance and the memory footprint.

\begin{table*}[h]
\footnotesize
\begin{center}
\caption{Different architecture configurations of DCNNs.}
\label{tab:structure}
\begin{tabular}{ccccc}
\toprule
\textbf{Model}  & \textbf{CIFAR-10}  & \textbf{CIFAR-100} &\textbf{Layer size } & \textbf{\# Parameters}\\ \midrule
CNN & 9.85\% & 34.26\% & 128 &1.\\ 
DCNN-32-6-3-2& 9.05\%  &  32.28\% & 128 & 1.\\ 
DCNN-16-6-3-1 & 9.16\% &  32.54\% & 256 &1.\\ 
DCNN-4-10-3-1 & 9.65\% &  33.57\% & 256 & 0.69\\ 
DCNN-128-4-3-2 & 8.58\% & 30.35\% & 128 &1.78 \\
\bottomrule
\end{tabular}
\end{center}
\end{table*}

\section{Conclusion}
We have proposed the doubly convolutional neural networks (DCNNs), which utilize a novel double convolution operation to provide an additional level of parameter sharing over CNNs. We show that DCNNs generalize standard CNNs, and relate to several recent proposals that explore parameter redundancy in CNNs. A DCNN can be easily implemented by modern deep learning libraries by reusing the efficient convolution module. DCNNs can be used to serve the dual purpose of 1) improving the classification accuracy as a regularized version of maxout networks, and 2) being parameter efficient by flexibly varying their architectures. In the extensive experiments on CIFAR-10, CIFAR-100, and ImageNet datasets, we have shown that DCNNs significantly improves over other architecture counterparts. In addition, we have shown that introducing the doubly convolutional layer to any layer of a CNN improves its performance. We have also experimented with various configurations of DCNNs, all of which are able to outperform the CNN counterpart with the same or fewer number of parameters.

\small
\bibliographystyle{unsrt}
\bibliography{nips_2016}

\end{document}